# Deep Features Analysis with Attention Networks

**Shipeng Xie, Da Chen, Rong Zhang, Hui Xue**

Alibaba Group Turing Lab

tangqiu.xsp, chen.cd, stone.zhangr, hui.xueh@alibaba-inc.com

## Abstract

Deep neural network models have recently draw lots of attention, as it consistently produce impressive results in many computer vision tasks such as image classification, object detection, *etc*. However, interpreting such model and show the reason why it performs quite well becomes a challenging question. In this paper, we propose a novel method to interpret the neural network models with attention mechanism. Inspired by the heatmap visualization, we analyze the relation between classification accuracy with the attention based heatmap. An improved attention based method is also included and illustrate that a better classifier can be interpreted by the attention based heatmap.

## Introduction

With the development of deep learning, various of deep neural networks consistently show their capabilities on different computer vision tasks. However, neural network based methods are hard to interpret and often treated as a blackbox. To better understand the network and the reason why it perform well, model interpretability and visualization of a neural network are increasingly crucial, as it can not only explain why the network can learn features to make each prediction, but also make it to be interpretable to diagnose the networks and optimize network architectures.

In this paper, we propose a new aspect to interpret the neural network models. The proposed method can visualize feature representations and diagnose networks to optimize the representation learning. It combines feature representation visualization of neural networks and attention mechanism. By showing the visualization of feature heatmap under different attention mechanism setup and their corresponding image classification accuracy, we can interpret the relation between them and better understand the key issues which affect the performance of the deep neural network based models.

## Related Work

Recently, various ways of networks interpretation have been proposed (Zhang, Wu, and Zhu 2017; Zhou et al. 2018; Fan 2017; Zhang and Zhu 2018; Zeiler and Fergus 2014).

As summarized in survey (Zhang and Zhu 2018), there are many different ways to explain the neural network. Visual interpretability can be classified into six different directions, such as visualization, diagnosis, building explainable models, *etc*. In this paper, we focus on visualization and diagnosis of CNN representations. The mainstream method of visualizing the model is to make use of the gradient to estimate the relation between the appearance of images with the unit score (Zeiler and Fergus 2014). Beyond visualization, One type of methods is to interpret the most salient portions of input image by creating a heatmap (Bach et al. 2015; Zhou et al. 2016). (Zhou et al. 2018) further analyze the network by introducing decomposition basis based on the heatmap and visualize the internal representations of a network, which can be treated as evidence of sensitivity of the most salient parts of images. (Zhang, Wu, and Zhu 2017) propose an interpretable covolutional neural network which is able to clarify knowledge representations in high convolutional layers of the CNN. In this paper, we apply the attention mechanism to analyze the relation between the heatmap and the classification accuracy.

Similar to the human perception process (Mnih et al. 2014), Attention mechanism is important for guiding bottom-up feed forward process with top information. The top-down attention mechanism has also been applied in image classification, such as sequential process (Mnih et al. 2014; Anne Hendricks et al. 2016; Xu et al. 2015), region proposal (Fu, Zheng, and Mei 2017; Xiao et al. 2015) and control gates, which capture different kinds of attention in a goal-driven way. Comparing to these methods, bottom-up top-down structure of feed forward attention (Wang et al. 2017; Newell, Yang, and Deng 2016; Fu, Zheng, and Mei 2017) has been applied to explicitly focus location of interest and explores relative information and achieves good results. In (Wang et al. 2017), mixed attention mechanism is built by stacking attention models multiple times, which separately generates attention masks. In (Fu, Zheng, and Mei 2017), a recurrent attention network is used to locate the discriminative area recurrently for better classification performance.

Attention based heatmap has been applied to benefit self-driving cars(steering control) (Kim and Canny 2017). Different from this work which follows the Encoder-Decoder technique to generate attention map to guide steering control, in this paper, we focus on visualizing the internal feature representation of a common image classification network, *i.e.*, Inception-v1 (Szegedy et al. 2015) and diagnosing the attention of the region of interest(ROI) for optimizing learning representative features.

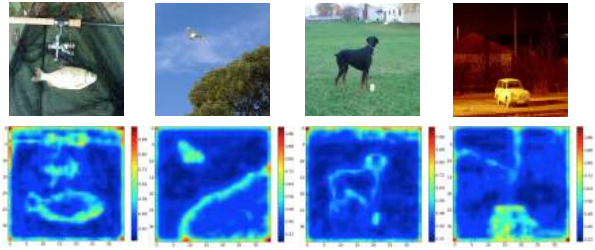

Figure 1: Example heatmaps: Top row are original images and bottom row are heatmaps.

## Method and Experiment

In this paper, we identify the evidence for image classification by creating a semantically interpretable heatmap of the internal features of a network. Examples are as shown in Figure 1. Noted that the most sensitive regions in the heatmap, indicating by warmest colour, are related to the most salient parts of input. It interprets the focus of a network that whether it takes the best attention on the relative ROI and makes use of context information. This is similar to attention mechanism, which is not only formulated as a selector to focus on the location of ROI but also used to enhance different representations of salient portions of input. Inspired by this observation, in this paper, we aim to interpret the network model by first visualizing the network with a heatmap, and then diagnosing the capability of a common image classifier by attention mechanism.

In order to interpret the evidence that a network applies to make a classification decision, we need visualization of the heatmaps of different layers in the network model. As different layers response to different intensity, the reason why the network make a confident decision can be analyzed by viewing the sensitivities in heatmaps of different layers. As shown in Figure 2, the heatmap of bottom layers with less semantic information interprets more attention on low-level features, while the heatmap of top layers with more semantic information shows more high-level attention on the most salient portions.

As studied in previous work about attention mechanism (Mnih et al. 2014; Xu et al. 2015; Anne Hendricks et al. 2016), the attention ability of a network for salient portions affects the capabilities of a network for making an accurate decision in consistent with the visualized heatmaps. If the network pays more attention on salient portions with ROI, the heatmaps show high sensitivity in the corresponding region and vice versa.

In this paper, we propose an attention based network for further comparison. In the proposed network, an unified bottom-up top-down structure is used to build sharable attention mechanism. Different scale of atttentions are created by generating different attention masks from the sharable structure. In a bit detail, two 1x1 convolutional layers are added from the structure, followed by sigmoid activation.

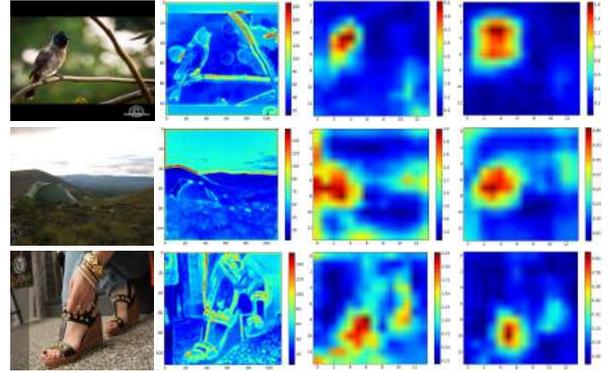

Figure 2: Heatmaps of different layers: First column are original images, second column are low-level features, third column are mid-level features and last column are high-level features.

|  | Inception-v1 | RAN-92 | Inception-v1-ours |
|---|---|---|---|
| Accuracy | 93.9% | 95% | 95.1% |

Table 1: Accuracy on CIFAR-10 of different image classification methods.

For simplicity, we visualize mid-level of feature maps for illustration. As shown in Figure 3, the first column is the original images. The second column is original heatmaps based on Inception-v1 (Szegedy et al. 2015) while the third column is heatmaps of a network with attention mechanism (Wang et al. 2017). Inspired by original heatmaps visualization, we enhance attention mechanism to optimize network by building a sharable structure. Attention mask is created with sigmoid activation function and then multiplied with original heatmap. The fourth column is heatmaps of the proposed network with better attention mechanism which we include a shareable attention mechanism.

As shown in Table 1, with attention mechanism, the image classifier can gain better performance. Noted that the result of our method is combining sharable attention mechanism and atrous convolution. As shown in Figure 3, we can observe that the attention-based method which can obtain most important part(warmer colour) of the ROI can get better result(Inception-v1 vs. RAN-92), even through RAN-92 got more 'noise' comparing to Inception-v1. Our method with a shareable attention network can obtain important features with less 'noise' and have the best performance. We can conclude that visualizing heatmap of a network shows whether it pays close attention on salient parts. Also, involving attention mechanism can benefit learning representative features of ROI for better focusing on salient portions. It is an interpretable way to compare original features to attention-focused features by creating heatmaps as visualization of feature representations.

We also design an experiment to verify the effects of different stages of involving attention mechanism for optimizing learning representations. In detail, we add attention mechanism into three different places of a network for comparison: early stage, middle stage and later stage respectively. For simplicity, we only add attention strategy for comparison, without atrous convolution which applied in the test method in Table 1. To compare the effects, we visualize heatmaps of feature representations in these three stages for verification. As shown in Figure 4, the sensitivity of heatmaps shows that adding attention mechanism at different stages brings different effects. It can be concluded from heatmaps in early stage that attention supervision information is critical for focusing on ROI at early stage of a network. As shown in second column of Figure 4, adding attention information at early stage can provide important attention with less noise. Additionally, we compare the accuracy of classification with attention mechanism at different stages. As shown in Table 2, it consistently shows improvements of adding attention mechanism at early stage of a network.

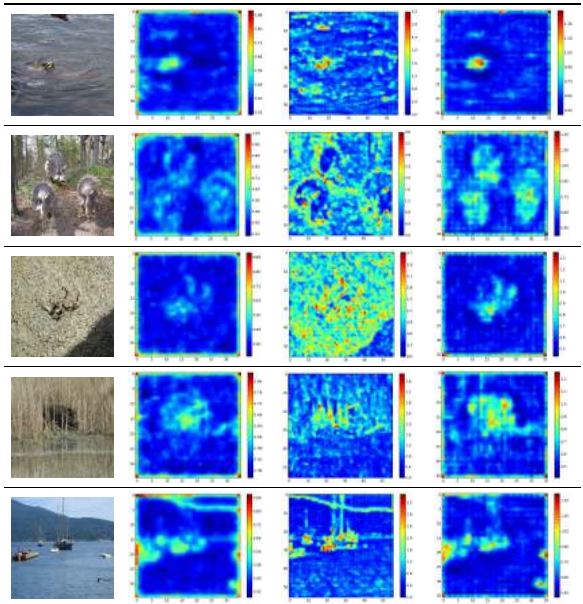

Figure 3: Examples of visualization of heatmaps with attention: First column are original images; second column are heatmaps based on Inception-v1; third column are heatmaps based on RAN-92; last column are heatmaps based on our method.

|  | Early Stage | Middle Stage | Later Stage |
|---|---|---|---|
| Accuracy | 94.71% | 94.55% | 94.23% |

Table 2: Accuracy on CIFAR-10 for adding attention mechanism on image classification model at different stages.

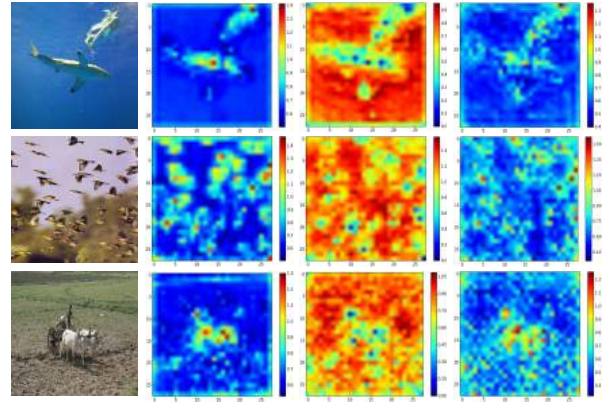

Figure 4: Effect of involving attention mechanism at different stages. First column are original images, while the rest three columns are adding attention mechanism at early, middle and later stage respectively.

## Conclusion and Discussion

In this paper, we visualize the internal feature representations of a network, *i.e.*, Inception-v1 (Szegedy et al. 2015) to interpret the capability of a network by creating heatmaps within different layers. Inspired by heatmap visualization, we bridge the connection between the attention mechanism and visualization of heatmap by creating attention masks. It guides to diagnose the focus of salient parts and optimize the ability of a network for learning representative features. By introducing a shareable attention network, we observe that adding attention mechanism in the early stage provide the best attention based heatmap which leads to the best performance in image classification task. We conclude that the attention based heatmap with best results should confidently have the most salient area(warmer colour) with less 'noise'. The work in this paper remains lots of future work, for example, we will try to exploit attention loss and reduce FLOPs of our method, *etc*.

## References

Anne Hendricks, L.; Venugopalan, S.; Rohrbach, M.; Mooney, R.; Saenko, K.; and Darrell, T. 2016. Deep compositional captioning: Describing novel object categories without paired training data. In *Proceedings of the IEEE Conference on Computer Vision and Pattern Recognition*, 1–10.

Bach, S.; Binder, A.; Montavon, G.; Klauschen, F.; Müller, K.-R.; and Samek, W. 2015. On pixel-wise explanations for non-linear classifier decisions by layer-wise relevance propagation. *PloS one* 10(7):e0130140.

Fan, L. 2017. Deep epitome for unravelling generalized hamming network: A fuzzy logic interpretation of deep learning. *arXiv preprint arXiv:1711.05397*.

Fu, J.; Zheng, H.; and Mei, T. 2017. Look closer to see better: Recurrent attention convolutional neural network for fine-grained image recognition. In *CVPR*, volume 2, 3.

Kim, J., and Canny, J. F. 2017. Interpretable learning for self-driving cars by visualizing causal attention. In *ICCV*, 2961–2969.

Mnih, V.; Heess, N.; Graves, A.; et al. 2014. Recurrent models of visual attention. In *Advances in neural information processing systems*, 2204–2212.

Newell, A.; Yang, K.; and Deng, J. 2016. Stacked hourglass networks for human pose estimation. In *European Conference on Computer Vision*, 483–499. Springer.

Szegedy, C.; Liu, W.; Jia, Y.; Sermanet, P.; Reed, S.; Anguelov, D.; Erhan, D.; Vanhoucke, V.; and Rabinovich, A. 2015. Going deeper with convolutions. In *Proceedings of the IEEE conference on computer vision and pattern recognition*, 1–9.

Wang, F.; Jiang, M.; Qian, C.; Yang, S.; Li, C.; Zhang, H.; Wang, X.; and Tang, X. 2017. Residual attention network for image classification. *arXiv preprint arXiv:1704.06904*.

Xiao, T.; Xu, Y.; Yang, K.; Zhang, J.; Peng, Y.; and Zhang, Z. 2015. The application of two-level attention models in deep convolutional neural network for fine-grained image classification. In *Proceedings of the IEEE Conference on Computer Vision and Pattern Recognition*, 842–850.

Xu, K.; Ba, J.; Kiros, R.; Cho, K.; Courville, A.; Salakhutdinov, R.; Zemel, R.; and Bengio, Y. 2015. Show, attend and tell: Neural image caption generation with visual attention. *arXiv preprint arXiv:1502.03044*.

Zeiler, M. D., and Fergus, R. 2014. Visualizing and understanding convolutional networks. In *European conference on computer vision*, 818–833. Springer.

Zhang, Q.-s., and Zhu, S.-C. 2018. Visual interpretability for deep learning: a survey. *Frontiers of Information Technology & Electronic Engineering* 19(1):27–39.

Zhang, Q.; Wu, Y. N.; and Zhu, S.-C. 2017. Interpretable convolutional neural networks. *arXiv preprint arXiv:1710.00935* 2(3):5.

Zhou, B.; Khosla, A.; Lapedriza, A.; Oliva, A.; and Torralba, A. 2016. Learning deep features for discriminative localization. In *Proceedings of the IEEE Conference on Computer Vision and Pattern Recognition*, 2921–2929.

Zhou, B.; Sun, Y.; Bau, D.; and Torralba, A. 2018. Interpretable basis decomposition for visual explanation. In *Proceedings of the European Conference on Computer Vision (ECCV)*, 119–134.